\pdfoutput=1
\documentclass{article}

\usepackage{microtype}
\usepackage{graphicx}
\usepackage{booktabs}
\usepackage{tabularx}
\usepackage{multirow}
\usepackage{float}
\usepackage{capt-of}
\usepackage{xurl}
\newcolumntype{L}{>{\raggedright\arraybackslash}X}
\usepackage{hyperref}

\usepackage{paperstyle}
\usepackage{amsmath}
\usepackage{amssymb}
\usepackage[capitalize,noabbrev]{cleveref}
\graphicspath{{figures/}}
\hypersetup{pdftitle={Same Pieces, Different Servers: A Tetris Benchmark for AI Agents as Served}, pdfauthor={Haochuan Wang}}

\newcommand{\tsc}[1]{\texttt{#1}}
\icmltitlerunning{Same Pieces, Different Servers: A Tetris Benchmark for AI Agents as Served}
\begin{document}

\twocolumn[
  \icmltitle{Same Pieces, Different Servers: A Tetris Benchmark for AI Agents as Served}
  \begin{icmlauthorlist}
    \icmlauthor{Haochuan Wang}{mit}
  \end{icmlauthorlist}
  \icmlaffiliation{mit}{Massachusetts Institute of Technology, Cambridge, MA, USA}
  \icmlcorrespondingauthor{Haochuan Wang}{hcw@mit.edu}
  \icmlkeywords{agents, LLM serving, decision models, evaluation, Tetris}
  \vskip 0.22in
  {\begin{center}
    \includegraphics[width=0.9\textwidth]{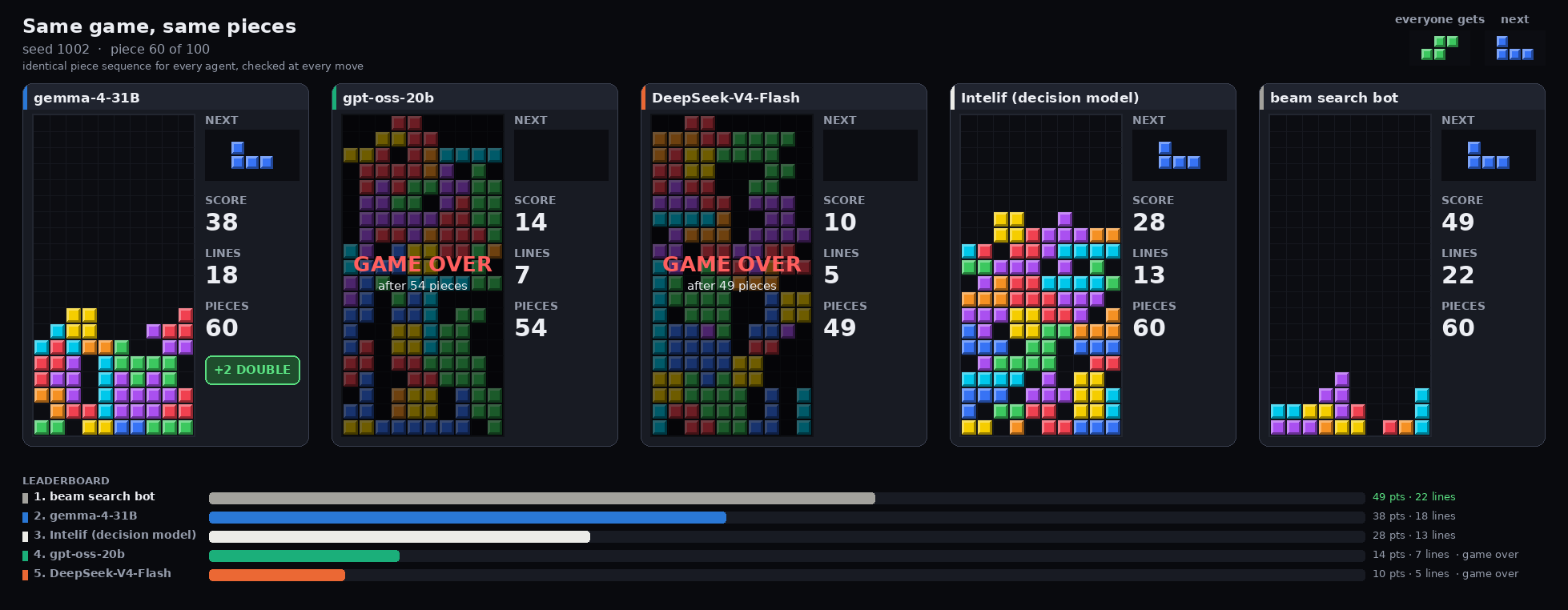}
    \captionof{figure}{\textbf{Same game, same pieces, five agents} (seed 1002, piece 60 of 100). Three open-weight language models served by Tensormesh and the Intelif decision model self-hosted on a CPU, replayed move by move from their logs, next to a beam-search bot, a classical planner that serves as the reference player (\cref{sec:method}). gpt-oss-20b and DeepSeek-V4-Flash have topped out; Intelif's stack is high, but it finishes all 100 pieces with 54 points, behind gemma-4-31B (76). Animations of more games are in the \tsc{gallery/} folder of the code repository (\cref{sec:intro}).}
    \label{fig:teaser}
  \end{center}}
  \vskip 0.15in
]

\printAffiliationsAndNotice{}
\setcounter{footnote}{1} 

\begin{abstract}
An agent meets a model as served: through an endpoint with a price card, a shared cache and other tenants, or on whatever hardware a self-hosted model runs. Benchmarks rank the weights. We introduce a Tetris benchmark that measures what agents get from models as served: every move is scored against an oracle, and every agent receives the same pieces. In five pre-specified experiments with nine open-weight models on one serverless provider, plus an open decision model self-hosted on a CPU, we find that price and size do not predict decision quality; that resending history costs almost nothing when cached input is free, would cost eleven to twelve times more if it were not, and makes play worse; that deployments keep between 4 and more than 64 agent contexts warm, in line with their throughput rather than the model's KV-cache size; that an agent's own long requests slow its slowest short decisions more than tenfold, which a simple admission rule cuts by 59\% without hurting play; and that the decision model plays mid-pack but takes 27\,s per move on a CPU, about 650 times longer than reported on a GPU. Architecture predicts some of what an agent sees; the deployment sets the rest.
\end{abstract}

\section{Introduction}
\label{sec:intro}

An agent built on a language model rarely runs the model itself. It calls an endpoint, and behind the endpoint sit an inference engine, a cache, a scheduler and a price list, all shared with other tenants. The agent adds its own habits: how much history it resends, how many requests it keeps in flight, and whether long and short requests share a queue. What the agent receives, a decision of some quality at some latency and price, comes from all of these together.

Evaluations mostly rank the weights. Agent and game benchmarks report task quality through an API and treat serving as fixed; serving systems are evaluated on servers their authors control, with quality held constant. For open-weight models the gap is sharpest: one checkpoint can be served by many providers, quantized, cached and priced differently. A newer kind of model widens it. Decision models such as the open Intelif do not generate text; they score a caller's options in one forward pass \citep{intelif}. Their appeal is speed, and speed is a property of where a model runs.

We measure what agents get from models as served, from the only place an agent stands: the client side. The instrument is Tetris, the classic video game \citep{algorta}. Every move can be scored against an oracle, every game replays exactly, and every agent can be given the same pieces (\cref{fig:teaser}). We study nine open-weight models served by one serverless provider, Tensormesh, and Intelif self-hosted on a 4-core CPU. Four findings stand out:
\begin{itemize}\setlength\itemsep{1pt}
  \item \textbf{Price and size do not buy decisions.} The most expensive model played worst; the best play came from a 117B model at \$0.015 per 100 moves.
  \item \textbf{Deployments keep 4 to more than 64 concurrent agent contexts warm.} The number follows the deployment's throughput, not the model's KV-cache size.
  \item \textbf{An agent's own long requests slow its short ones.} The 95th-percentile time to a move grows from 1\,s to 11\,s; deferring the long requests cuts it by 59\% at the same cost, on held-out data.
  \item \textbf{A decision model is fast only where it is served.} Intelif played fifth of ten agents on the same games, always legally, but took 27\,s per move on our CPU against about 40\,ms reported on one GPU.
\end{itemize}
Our contributions are a measurement design that separates what an agent controls, observes and cannot see (\cref{sec:setup}); five pre-specified experiments that link deployment properties to cost, latency and decision quality, read against each model's published architecture (\cref{sec:models,sec:method,sec:results}); and an open Tetris benchmark that replays every game and plays the same games against language models and decision models alike.\footnote{Code, raw logs and animations: \url{https://github.com/KevinChunye/Diffusion-Tetris}.}

\section{What an Agent Can See}
\label{sec:setup}

\Cref{tab:setup} lists the layers between a Tetris agent and the weights, what our agent varies at each layer, what it can observe, and what stays hidden. For the served models we vary only the top two layers and the choice of model. Everything we say about the engine is inferred from timing and reported token counts, so we state it as a hypothesis consistent with the data, not as a measurement.

\begin{table*}[t]
\caption{The measurement setup, layer by layer. E1--E4 call Tensormesh's serverless API (vLLM with LMCache); E5 runs a decision model on our own hardware. Engine internals are never observed directly.}
\label{tab:setup}
\centering\small
\begin{tabularx}{\textwidth}{@{}>{\raggedright\arraybackslash}p{2.6cm}LLL@{}}
\toprule
Layer & What the agent varies (experiment) & What it observes & What stays hidden \\
\midrule
Task: Tetris & nothing: every agent gets the same seeded pieces & board, current and next piece, legal placements; oracle regret per move & nothing (fully observed) \\
Agent harness & model (E1), history resent (E2), concurrent agents (E3), request mix and admission (E4) & its own queue and timing & nothing \\
Serverless API & output limits, JSON schema, priority hints (E4) & latency; reported prompt, cached and output tokens; refusals (HTTP 429); the public price card & billing records, routing, replica count, other tenants \\
Inference engine & nothing & only indirectly: cache hits, prefill throughput & batching, scheduler, KV memory pool, eviction and offloading \\
Weights & which served model id (E1--E4) & model card and config: parameters, attention and KV design & the exact checkpoint and quantization behind an id \\
Decision model & hardware (E5) & the probability of every option; exact compute time & nothing (self-hosted) \\
\bottomrule
\end{tabularx}
\end{table*}

\section{Related Work}

\textbf{Serving systems.} Continuous batching \citep{orca}, paged KV memory in vLLM \citep{vllm}, chunked prefill \citep{sarathi} and prefill/decode disaggregation \citep{distserve,splitwise} make engines fast; KV reuse across requests is handled by RadixAttention \citep{sglang}, multi-turn caching \citep{cachedattn}, KV-centric designs \citep{mooncake} and cache layers such as LMCache \citep{lmcache}. Agent-aware engines schedule whole programs \citep{autellix}. This work assumes control of the server; we ask what of it reaches a tenant. Client-side scheduling for black-box APIs has been studied in simulation \citep{yuan}; we test its simplest form on a live endpoint, with open-loop load \citep{schroeder} and tail metrics \citep{tail}.

\textbf{Benchmarks and audits.} AgentBench \citep{agentbench}, BALROG \citep{balrog} and lmgame-Bench \citep{lmgame}, which includes Tetris, score decisions through an API; HELM \citep{helm} adds efficiency metrics, and MLPerf Inference \citep{mlperf} measures serving stacks at fixed accuracy. Audits show that deployments differ: prompt caches shared across users \citep{gu}, undetectable model substitution \citep{cai}, and vendor differences for one open model \citep{k2}. We measure what such differences cost an agent.

\textbf{Decision models.} A decision model returns a probability for each of a caller's options instead of generating text, so one forward pass scores every option. Intelif is an open example: Qwen3-4B \citep{qwen3} with a LoRA adapter \citep{lora} and a linear scorer that reads one anchor token per option \citep{intelif}. Its authors report 31.77 on the third-party Decision Index \citep{dindex} at a median of 16.5\,ms on one RTX PRO 6000 GPU, and about 40\,ms per Tetris move. We test it where an agent without a GPU would run it.

\section{Models, and What Their Architecture Predicts}
\label{sec:models}

The provider serves nine open-weight models (\cref{tab:models}). If the weights alone decided what an agent experiences, two published quantities should matter most. The first is the KV cache one agent context occupies, which bounds how many contexts a given memory can keep; across these models it differs by a factor of 64. The second is the number of active parameters, which sets the compute to read a prompt, about two FLOPs per active parameter per token plus attention; it differs by a factor of 11.

\begin{table*}[t]
\caption{The models. Parameters and attention design are from each model's report, card or config. KV/ctx is our estimate of the 16-bit KV cache (plus the fixed recurrent state of Gated DeltaNet layers) that one 5.5k-token agent context needs, computed from the published config; real engines may store less (e.g., FP8). Prices are the provider's list prices per million input and output tokens; the provider states that cached input is free on every model. $^a$Active parameters from the MiniMax-M2 card (same architecture). $^b$Not on the price page, which lists the previous version (Qwen3.6-27B, Kimi K2.6, GLM-5.1) at these prices.}
\label{tab:models}
\centering\footnotesize
\begin{tabularx}{\textwidth}{@{}lrLrrl@{}}
\toprule
Model & Total / active & Attention and KV design & KV/ctx & \$/M in, out & Source \\
\midrule
gpt-oss-20b & 21B / 3.6B & alternating full and 128-token window & 132 MB & 0.07, 0.28 & \citet{gptoss} \\
gpt-oss-120b & 117B / 5.1B & same pattern, 36 layers & 198 MB & 0.15, 0.60 & \citet{gptoss} \\
gemma-4-31B & 31B dense & 5 sliding (1,024) : 1 global layer & 1,015 MB & 0.14, 0.56 & \citet{gemma4} \\
DeepSeek-V4-Flash & 284B / 13B & compressed sparse attention; FP8 cache & 21 MB & 0.14, 0.28 & \citet{dsv4} \\
MiniMax-M2.5 & 229B / 10B$^a$ & full attention in all 62 layers & 1,335 MB & 0.30, 1.20 & \citet{minimax} \\
Qwen3.5-397B & 397B / 17B & 3 Gated DeltaNet : 1 full, 60 layers & 342 MB & 0.60, 3.60 & \citet{qwen35} \\
Qwen3.8-27B & 27B dense & 3 Gated DeltaNet : 1 full, 64 layers & 489 MB & 0.32, 3.20$^b$ & model card \\
Kimi-K2.7 & 1T / 32B & latent attention; INT4 experts & 370 MB & 0.96, 4.00$^b$ & \citet{kimi} \\
GLM-5.2 (NVFP4) & 744B / 40B & latent + sparse attention & 525 MB & 1.40, 4.40$^b$ & \citet{glm5} \\
\midrule
Intelif (self-hosted) & 4.0B dense & Qwen3-4B + LoRA + linear scorer & 775 MB & -- & \citet{intelif} \\
\bottomrule
\end{tabularx}
\end{table*}

The models shrink the KV cache in different ways. Grouped-query attention shares key--value heads across query heads \citep{gqa}; sliding-window layers keep only the last 128 (gpt-oss) or 1,024 (gemma) tokens; latent attention stores one compressed vector per token, which cut the KV cache by 93\% in DeepSeek-V2 \citep{dsv2}; DeepSeek-V4 compresses further and keeps an FP8 cache. Sparse attention reduces attention compute rather than memory, although its indexer still scales quadratically \citep{dsv32,indexcache}. The Qwen models replace three of every four attention layers with Gated DeltaNet \citep{gdn}, a recurrent layer with a fixed-size state. Such hybrids are harder to prefix-cache: vLLM stores their state only at block boundaries, a mode it marks experimental \citep{vllmhybrid}. MiniMax kept full attention in every layer, citing high cache-hit rates in real conversations as one reason \citep{minimax}. These designs give three predictions an agent can check from outside: if memory per context limited caching, capacity should fall as KV per context grows (P1); cold prefill time should grow with active parameters (P2); and the hybrid models should report only part of a repeated prefix as cached (P3).

\section{Method}
\label{sec:method}

\textbf{Task.} The board is $10\times20$ with one piece of preview, and pieces are drawn uniformly at random from a generator seeded per game. The piece sequence depends only on the seed, never on how much an agent simulates, so all agents face identical games. Games stop at 100 pieces or when the stack tops out.

\textbf{Language-model agents (E1--E4).} Each move, the agent receives the board, the current and next piece, and the legal placements, each annotated with its simulated outcome (lines cleared, resulting height and holes). It answers with a JSON object naming one placement; in E1 and E2 a JSON schema restricts the output to legal placements. gpt-oss models run at low reasoning effort (gpt-oss-120b also at medium, as a separate configuration); gemma, Qwen, GLM and Kimi have thinking switched off through their chat template, and DeepSeek-V4-Flash answers directly by default. MiniMax-M2.5 cannot switch reasoning off, so it appears only in E3.

\textbf{Decision-model agents (E5).} A decision model receives the same board and pieces; each legal placement is one option, described in words by its simulated outcome, following Intelif's own Tetris example, and the agent plays the most probable option. Intelif runs with its released weights on a 4-core server CPU without a GPU, the hardware an agent without an accelerator would have.

\textbf{Reference players.} Two programs anchor the score scale. A random player picks a uniformly random legal placement and defines 0. The beam-search bot defines 1. It is a classical planner with no learned parameters: for every legal placement of the current piece it simulates the result with the game engine and scores the board with a fixed formula that rewards cleared lines and penalizes holes, height and bumpiness; it keeps the 16 best boards, extends each with the previewed next piece and then with one further piece that it samples, because it cannot see it, and plays the first move of the best three-move plan. It sees exactly what the agents see (board, current and next piece, legal placements) and receives the same pieces. It differs in how it decides. The language-model agents get a one-move version of the same simulation as annotations and must choose in a single call; the decision model scores the same list of placements in one forward pass; neither looks further ahead. The bot searches ahead instead, needs no model, network or accelerator, and its play does not depend on any deployment, which makes it a fixed yardstick rather than a competitor. The oracle used for regret is a separate lookahead (rollouts plus a narrow beam search) that only grades moves.

\textbf{Outcomes.} Quality is the game score relative to a beam-search bot on the same seeds, pooled over seeds as $(\sum \text{score} - \sum \text{random}) / (\sum \text{beam} - \sum \text{random})$, and the per-move regret against an approximate oracle that scores every legal placement by rollouts and a narrow beam search with common random numbers. Latency is wall time per decision measured by the client. Cost to the agent is computed from list prices and the token counts the API reports. The provider states that cached input is free on every model, but four model cards show no cached price, so where it matters we also show the cost with cached input billed at the full input price.

\textbf{Capacity probe (E3).} $N$ agents each send a unique context of about 5.5k tokens (a random nonce in the system prompt followed by eight turns of a game) with a one-token output, stay idle for 30\,s, then send the same context plus one turn. An agent counts as warm when the second request reports at least half of the first one's prompt as cached and both requests succeeded. GLM-5.2 reports no cached tokens, so its hits are inferred from latency (second request faster than 0.35 of its cold median); this rule agrees with reported hits on 92\% of agents on the eight models that report them. $N$ runs from 1 to 64 in randomized order, with one to five repetitions per model. Aggregate prefill throughput is $N$ times the context length divided by the slowest first request of the round.

\textbf{Design and statistics.} Hypotheses, metrics and decision rules for every experiment were fixed before its data were collected (\cref{app:experiments}). Games are the independent unit: intervals for scores and paired differences come from 2,000 bootstrap resamples of seeds, and rank correlations are bootstrapped the same way. Hit rates carry Wilson 95\% intervals. E4 uses a two-level paired bootstrap over time blocks and requests. Experiments that measure the cache or latency ran without our other traffic.

\section{Results}
\label{sec:results}

\subsection{Price and size do not buy decisions (E1)}

\begin{figure*}[t]
  \centering
  \includegraphics[width=\textwidth]{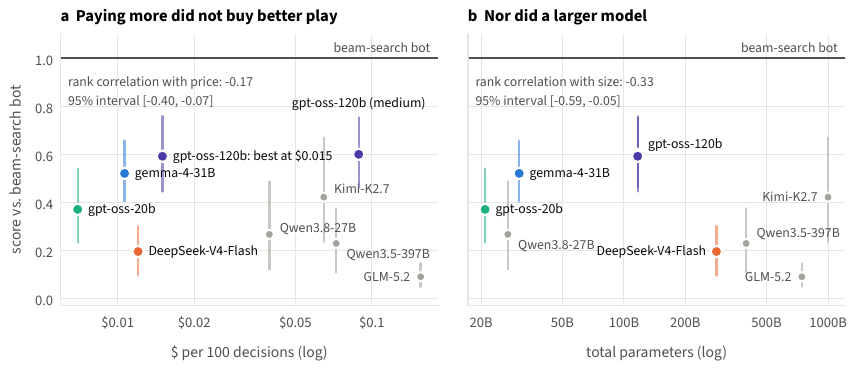}
  \caption{\textbf{Neither price nor size predicts how well a model plays.} Score relative to a beam-search bot on the same seeds (pooled over ten seeds; bars: 95\% bootstrap intervals over seeds) against the estimated price per 100 decisions (a) and total parameters (b), nine configurations served by Tensormesh, no history, output constrained to legal placements. The four models used throughout the paper are colored; the others are gray.}
  \label{fig:ladder}
\end{figure*}

\Cref{fig:ladder} places nine configurations by price, size and how well they play, over ten seeds. The best play came from gpt-oss-120b, which reached 0.59 of the beam-search bot's score at low reasoning effort for \$0.015 per 100 decisions and 1.5\,s per move; medium effort reached 0.60 at six times the price and eight times the latency. gemma-4-31B followed at 0.52 for \$0.011. The most expensive model, GLM-5.2 at \$0.156 per 100 decisions, played worst (0.09), and the largest, Qwen3.5-397B, reached 0.23. Across the nine configurations the rank correlation between score and price was $-0.17$ (95\% bootstrap interval $-0.40$ to $-0.07$), with total parameters $-0.33$ ($-0.59$ to $-0.05$) and with active parameters $-0.41$ ($-0.63$ to $-0.23$), meeting our pre-specified criterion that neither price nor size shows a positive association. Per-move oracle regret orders the configurations almost identically, from 0.49 for gpt-oss-120b to 2.9 for GLM-5.2. A smaller, three-seed run of the same configurations ordered them quite differently (gemma 0.88, Kimi 0.85), a reminder that game benchmarks with few seeds rank noisily.

\subsection{What history costs, and what it does to play (E2)}

\begin{figure*}[t]
  \centering
  \includegraphics[width=\textwidth]{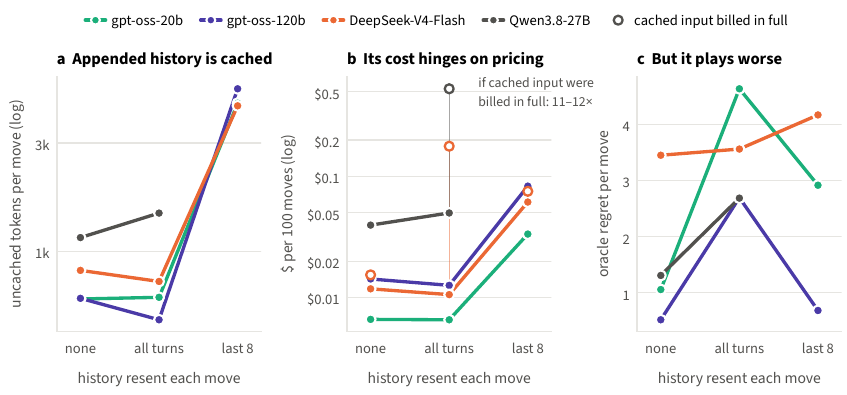}
  \caption{\textbf{Appending history is cheap only because the cache absorbs it, and it plays worse.} Three history policies on four models, 6--7 paired seeds each. (a) Prompt tokens per move not served from cache. (b) Estimated cost under the provider's stated pricing (cached input free); hollow markers show the cost if cached input were billed at the full input price, which matters for the two models whose cards list no cached price. (c) Oracle regret per move.}
  \label{fig:memory}
\end{figure*}

Agents carry memory by resending history. We compared three policies on four models: no history, appending every turn, and keeping the last eight turns (\cref{fig:memory}). Appending lets the cache do its job: the prompt grows past 10k tokens, yet 90--97\% of it is served from cache, so the tokens actually processed per move stay at 500--740 on three models, about the same as with no history (620--830). The eight-turn window is the expensive choice: dropping the oldest turn rewrites the prefix right after the system prompt, so every move re-processes 4.4k--5.2k tokens, 5.7 to 10 times as many as appending (paired over seeds; 95\% intervals within about 10\%). Qwen3.8 shows the hybrid cost of \cref{sec:capacity} in a real game: appending leaves about 1,475 tokens per move uncached, twice the other models, and its 1.15k-token prompt without history is never cached at all, because it is shorter than one cache block.

Who pays for the cache decides whether appending is cheap. Under the provider's stated policy, appending costs 0.89 to 1.27 times as much as no history on every model. If cached input were billed at the full price, as two of the four model cards would literally suggest, appending would cost 11 times as much on DeepSeek-V4-Flash (95\% CI 9.5--12.6) and 12 times on Qwen3.8 (8.9--15.2), while staying at 0.9--1.0 on the gpt-oss models. Quality points the other way. The board is fully visible, so history adds no information, and appending lowered every model's score relative to the bot: gpt-oss-20b from 0.40 to 0.03, gpt-oss-120b from 0.60 to 0.24, Qwen3.8 from 0.39 to 0.13 and DeepSeek from 0.20 to 0.11. Per-move regret rose by 3.6 on gpt-oss-20b (95\% CI 2.9--4.3, seven paired seeds) and by 2.2 on gpt-oss-120b (1.9--2.7); on Qwen3.8 ($+1.6$, $-0.5$ to $3.5$) and on DeepSeek, which already played poorly without history ($+0.03$, $-1.1$ to $1.4$), the change was not distinguishable from zero. The memory that is cheapest to serve is the one that plays worst, consistent with evidence that long, repetitive context degrades the use of its relevant part \citep{lost}.

\subsection{How many agents a deployment keeps warm (E3)}
\label{sec:capacity}

\begin{figure*}[t]
  \centering
  \includegraphics[width=\textwidth]{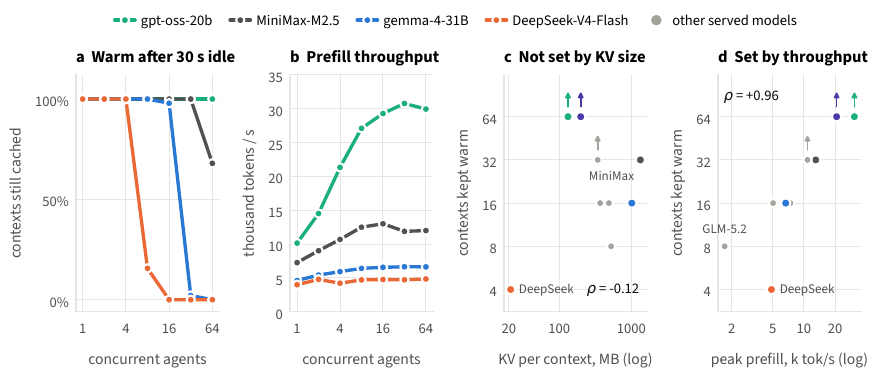}
  \caption{\textbf{How many agent contexts a deployment keeps warm follows its throughput, not the model's KV-cache size.} $N$ agents each send a unique 5.5k-token context, wait 30\,s and send it again. (a) Share still served from cache and (b) aggregate prefill throughput, for the four models whose curves show the pattern most clearly (all nine in \cref{fig:capacityall}). (c, d) Capacity, the largest $N$ with at least 90\% of contexts still cached, against the KV cache one context needs and against the deployment's peak aggregate prefill throughput; arrows mark capacities at the largest $N$ tested; $\rho$ is Spearman's rank correlation over nine models.}
  \label{fig:capacity}
\end{figure*}

Every deployment cached a repeated 5.5k-token context when one agent used it. How many agents it could serve that way differed by more than an order of magnitude (\cref{fig:capacity}; all nine models in \cref{fig:capacityall}). gpt-oss-20b and gpt-oss-120b kept all 64 concurrent contexts warm after 30\,s of idleness, the most we tested. MiniMax-M2.5 kept 32 and Qwen3.5-397B all 32 it was tested with; gemma-4-31B, Qwen3.8-27B and Kimi-K2.7 kept 16; GLM-5.2 kept 8; DeepSeek-V4-Flash kept 4. Beyond capacity the loss is abrupt: gemma kept 98\% of 16 contexts but 2\% of 32, and an evicted context took 12 to 19 times longer to serve again (2.3\,s instead of 0.19\,s on gemma; 6.3\,s instead of 0.33\,s on DeepSeek). Past capacity most deployments also refused requests: at 32 or 64 agents, 62\% of DeepSeek's agents and 54\% of gemma's had a request refused with HTTP 429, while the gpt-oss deployments refused none.

The architecture does not explain the ranking (P1). The model with the smallest KV cache per context, DeepSeek-V4-Flash at an estimated 21\,MB, keeps the fewest contexts, and the one with the largest, MiniMax-M2.5 at 1.3\,GB, keeps 32; across the nine models the rank correlation between capacity and KV size is $-0.12$. What capacity does track is how much prefill work a deployment can do at once ($\rho = +0.96$). gpt-oss-20b's aggregate throughput grew from 10k to 31k tokens per second as agents were added, a sign of many parallel replicas, while DeepSeek stayed near 4.8k and gemma near 6.6k, as if requests were processed by a single replica in turn. Measured in tokens, DeepSeek's capacity is the same at two context lengths: with 15k-token contexts it kept one but not four (between 15k and 60k tokens), with 5.5k-token contexts four but not eight (between 22k and 44k). Even 44k tokens of DeepSeek's compact cache is about 150\,MB by our estimate, far below the KV memory of one GPU, so the limit is a property of how much cache and hardware the deployment allots, or of how many other tenants share it, not of the model.

Two architectural predictions do hold. Cold prefill took longer on models with more active parameters (P2; Spearman $+0.82$ across the nine models), from 0.54\,s on gpt-oss-20b to 1.5\,s on Kimi and 3.6\,s on GLM-5.2. And the two Gated DeltaNet hybrids reported only part of a repeated prefix as cached (P3): Qwen3.5 always 5,280 of 5,684 tokens (93\%) and Qwen3.8 always 4,704 (83\%), exactly the prompt rounded down to a multiple of a large block, as block-aligned caching of the recurrent state predicts. The other models cache all but at most the last 140 tokens. A Qwen3.8 agent therefore re-pays up to about 1.5k tokens of prefill on every turn.

\begin{figure}[!t]
  \centering
  \includegraphics[width=\columnwidth]{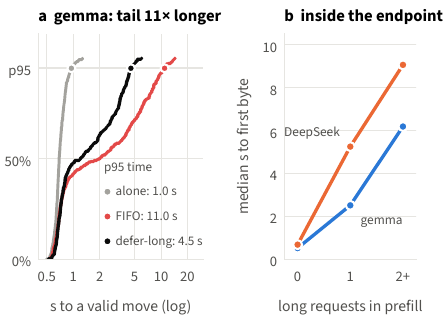}
  \caption{\textbf{Long requests stretch the tail of short decisions, inside the endpoint.} Short (about 1k-token) decisions sent alone or mixed with long (about 19k-token) requests at identical arrival times, ten time blocks per model. (a) Time from scheduled arrival to a valid move on gemma-4-31B, with each policy's 95th percentile. (b) Endpoint time to first byte of a short request under FIFO, by the number of long requests still waiting for their own first byte when it was sent.}
  \label{fig:interference}
\end{figure}

\begin{figure*}[t]
  \centering
  \includegraphics[width=\textwidth]{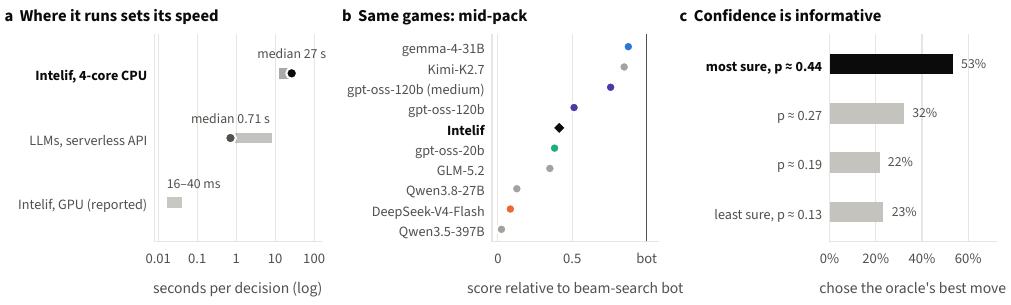}
  \caption{\textbf{A decision model is fast only where it is served.} (a) Seconds per decision (bar: range; dot: median) for Intelif on a 4-core CPU and for the nine language-model configurations through the serverless API, against the range its authors report on one GPU (gray). (b) Score relative to the beam-search bot on the same three games (seeds 1000--1002). (c) How often Intelif chose the oracle's best move, by quarter of the probability it gave its choice.}
  \label{fig:decision}
\end{figure*}

\subsection{An agent's own long requests slow its short ones (E4)}

Agents mix traffic: short per-move decisions and, now and then, long requests that summarize or plan over a large context. We replayed the same frozen requests, at the same arrival times, under four client policies: short requests alone, everything in one first-in-first-out (FIFO) queue, the same queue with vLLM-style priority hints, and defer-long, which sends a long request only when no short one is waiting or in flight, keeps at most one long request outstanding, and stops deferring after 6\,s. The requests come from a held-out corpus, with hypotheses and margins fixed in advance.

Mixing in long requests raised the 95th-percentile time to a valid move from 0.95\,s to 11.0\,s on gemma-4-31B ($+1{,}052\%$, 95\% CI $+689\%$ to $+1{,}386\%$) and from 1.8\,s to 22.4\,s on DeepSeek-V4-Flash ($+1{,}123\%$, CI $+372\%$ to unbounded), confirming the first hypothesis on both (\cref{fig:interference}). Most of the delay arises inside the endpoint, not in our queue: a short request's median time to first byte grew from 0.5\,s to 2.5\,s with one long request still in prefill and to 6.2\,s with two on gemma (0.7, 5.3 and 9.1\,s on DeepSeek), while a plain network probe moved by less than 0.2\,s. Priority hints, which the endpoint accepts, changed nothing measurable (gemma $-2\%$, CI $-16\%$ to $+5\%$). Defer-long cut gemma's tail by 59\% (CI 45\% to 66\%) and raised the rate of useful decisions by 24\% at the same cost, passing every pre-specified non-inferiority check, including the per-move regret margin. On DeepSeek, defer-long cut the tail by 46\% and raised useful decisions by 23\%, but about 4\% of short decisions never produced a valid move under mixed load, so the tail interval is unbounded and that hypothesis remains inconclusive. Defer-long cannot help short requests that arrive while a long one is already in prefill, and it makes long requests wait longer.

\subsection{Decision models as served (E5)}

Intelif answered every move with a legal placement and a probability for each option, so no output had to be parsed, retried or replaced. Over three games it reached 0.41 of the beam-search bot's score, fifth of the ten agents that played these seeds, between gpt-oss-120b at low effort (0.51) and gpt-oss-20b (0.38), although Tetris is not among its training tasks (\cref{fig:decision}b). Its per-move regret, 1.28, fell between the stronger language models (0.38--0.85) and the weaker ones (2.1--3.7). Its speed is where serving matters most. On the 4-core CPU each move took a median of 27\,s, growing linearly with the prompt at about 28\,ms per input token. Its authors report about 40\,ms per Tetris move and a median of 16.5\,ms on their benchmark, both on one GPU, roughly 650 times faster with the same weights, while the language models answered in a median of 0.6--0.7\,s through the serverless API (\cref{fig:decision}a). Its probabilities carry information at the top end: on the quarter of moves where it was most sure, it picked the oracle's best move 53\% of the time and its regret was 0.30, against 22--32\% in the other quarters (\cref{fig:decision}c), so an agent could use them to decide when to fall back to search or a larger model.

\section{Discussion}

\textbf{What the weights predict, and what they do not.} Published architecture predicted two things an agent can see from outside: how long a cold prompt takes (active parameters) and how much of a repeated prompt a hybrid model can reuse (block-aligned caching of its recurrent state). It did not predict the outcome that matters most for an agent fleet, how many agents a deployment keeps warm, which followed the hardware behind the deployment rather than the compactness of the model's KV cache. Neither did price or size predict how well a model played: the most expensive model played worst, and gpt-oss-120b at low reasoning effort matched its own medium-effort play at a sixth of the price. A leaderboard row reports none of this.

\textbf{What to report.} An evaluation of an agent on a served model should state the provider and model identifier, the date, the full price card including cached input and how it was applied, the cache behavior the API reports (granularity and hit rate at the agent's concurrency), refusals under load, the agent's memory and admission policy, and, for self-hosted models, the hardware and numeric precision. Each factor we varied or measured changed an outcome here by a factor of two or more.

\textbf{Decision models.} A decision model removes generation, parsing and invalid outputs from the loop: Intelif always returned a legal move, with probabilities informative enough to decide when to escalate. Its selling point, speed, belongs to the hardware: on the CPU an agent without a GPU would use, it was about 40 times slower than a far larger language model behind a serverless API. Comparisons between decision models and language models should state where each one ran.

\section{Limitations}

All API measurements come from one provider over two days, on a shared platform whose other traffic we cannot see; they describe these deployments at that time, not the models in general. Model identifiers do not prove which checkpoint or quantization was served. Capacity is measured with one context length, one idle gap and one to five repetitions per model, and refusals at high load also reduce the number of contexts that reached the cache. Costs are estimates from list prices and reported token counts. The oracle is approximate. Tetris has a fully visible state, which is likely why history did not help; tasks with hidden state may reward memory. The decision-model comparison covers three games with the prompt format of Intelif's own example; other prompts or hardware may change it.

\section{Conclusion}

For an agent, the product is not the weights but the model as served: behind a price card and a cache, next to other tenants, on whatever hardware runs it, and driven by the agent's own habits. Measured from the client, these factors changed what the agent paid by up to an order of magnitude, its tail latency by more than an order of magnitude, how many agents could share a deployment by a factor of sixteen, and a decision model's speed by roughly 650 times; price and size did not predict how well an agent played. Published architecture explained part of this, the deployment the rest. Evaluations of open-weight and decision models should report the serving context alongside the score.

\begingroup\raggedright
\bibliography{references}
\endgroup
\bibliographystyle{paperstyle}

\newpage
\appendix
\onecolumn
\raggedbottom

\section{Experimental Design}
\label{app:experiments}

\begin{table}[H]
\caption{The five experiments. Each was specified, with its hypotheses, metrics and decision rules, before its data were collected.}
\label{tab:experiments}
\centering\small
\begin{tabularx}{\textwidth}{@{}lLL@{}}
\toprule
Experiment & Design & Main measures \\
\midrule
E1 model ladder & 9 configurations $\times$ 10 seeds, up to 100 pieces; no history; output constrained to legal placements & score relative to the bot, oracle regret, price and latency per decision \\
E2 history policy & 4 models $\times$ 3 history policies (Qwen3.8: 2) $\times$ 9 seeds & uncached tokens and cost per move, oracle regret \\
E3 cache capacity & 9 models $\times$ $N \in \{1, \dots, 64\}$ concurrent agents $\times$ 1--5 repetitions & share of contexts still cached after 30\,s idle, refusals \\
E4 interference & 2 models $\times$ 10 time blocks $\times$ 4 client policies; held-out request corpus & 95th-percentile time to a valid move, useful decisions per second \\
E5 decision model & Intelif on a 4-core CPU, 3 games (seeds 1000--1002) & score relative to the bot, oracle regret, latency \\
\bottomrule
\end{tabularx}
\end{table}

\textbf{Exclusions.} E3 excludes rounds that overlapped in time with another deployment's large round, which drew refusals across the provider's gateway. E2 compares policies on the 6--7 seeds per model on which every policy completed its game. Scores relative to the bot are pooled over seeds rather than averaged per seed, because a per-seed ratio is unstable on seeds where the bot itself scores little.

\section{Cache Capacity of Every Model}
\label{app:capacity}

\begin{figure}[H]
  \centering
  \includegraphics{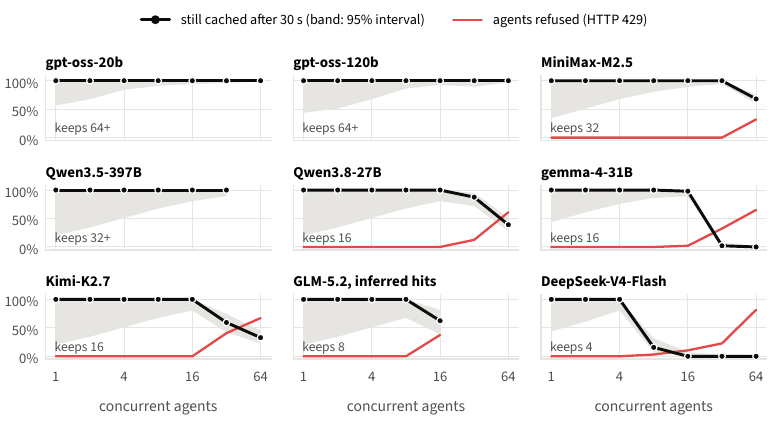}
  \caption{Cache capacity of every served model, ordered by capacity. Each panel shows the share of agents whose context was still served from cache after 30\,s (line; band: Wilson 95\% interval over agents) and the share of agents with a refused request (red), against $N$ concurrent agents with 5.5k-token contexts. One to five rounds per point. GLM-5.2 reports no cached tokens, so its hits are inferred from latency, a rule that agrees with reported hits on 92\% of agents on the eight models that report them. Kimi-K2.7, Qwen3.8, Qwen3.5 and GLM-5.2 were measured with one repetition each, so their intervals are wide.}
  \label{fig:capacityall}
\end{figure}

\section{The Decision-Model Agent}
\label{app:decision}

Each move is one multiple-choice question in Intelif's request format. The state holds the board as 20 rows of \tsc{\#} and \tsc{.}, the current piece and the next piece. Each legal placement is one option, keyed by rotation and column and described by its simulated outcome, following Intelif's own Tetris example, for instance ``rotation 0, column 3: clears no lines, no new holes, raises the stack by 2, surface gets bumpier''. The instruction reads ``You are playing Tetris. Where should the T piece land? Clear lines, avoid holes and keep the stack low and flat.'' Positions had 5 to 34 options. The language-model agents received the same facts as numbers, so \cref{fig:decision} compares agents as each is meant to be used, not the two kinds of model under identical inputs. Play is deterministic: replaying a game reproduced the same moves with identical probabilities.

\end{document}